\documentclass[11pt]{article}
\usepackage{acl2015}
\usepackage{times}
\usepackage{url}
\usepackage{latexsym}
\usepackage{natbib}
\usepackage{graphicx}
\usepackage{hyperref}
\usepackage{tabularx}
\usepackage{adjustbox}
\usepackage{multirow}
\usepackage{float}

\title{Exploring Hate Speech Detection with HateXplain and BERT}

\author{Aryan Mehra \\
  MS in CS \\
  Carnegie Mellon University\\
  { aryanm@andrew.cmu.edu} \\\And
  Arvind Subramaniam\\
  MS in ECE \\
  Carnegie Mellon University \\
  {arvindsu@andrew.cmu.edu} \\\And
  Sayani Kundu \\
  MCDS \\
  Carnegie Mellon University \\
  {sayanik@andrew.cmu.edu} \\}

\date{8 October, 2021}

\begin{document}
\maketitle
\begin{abstract}

Hate Speech takes many forms to target communities with derogatory comments, and takes humanity a step back in societal progress. HateXplain is a recently published and first dataset to use annotated spans in the form of 'rationales', along with speech classification categories and targeted communities to make the classification more human-like, explainable, accurate and less biased. We tune BERT to perform this task in the form of rationales and class prediction, and compare our performance on different metrics spanning across accuracy, explainability and bias. Our novelty is threefold - Firstly, we experiment with the amalgamated rationale-class loss with different importance $\lambda$ values. Secondly, we experiment extensively with the ground truth attention values for the 'rationales'. With the introduction of conservative and lenient attentions, we compare performance of the model on HateXplain and test our hypothesis. Thirdly, in order to improve the unintended bias in our models, we use masking of the target community words and note the improvement in bias and explainability metrics. Overall, we are successful in achieving model explanability, bias removal and several incremental improvements on the original BERT implementation. Reference to our code  \href{https://github.com/sayani-kundu/11711-HateXplain}{Github Repo}

%   Hate Speech targeted towards certain communities propagates bias and inequality in society, and takes humanity a step back in societal progress. A recently published dataset for hate speech detection - HateXplain, is the first dataset to use annotated spans in the form of 'rationales', along with speech classification categories and targeted communities to make the classification more human-like, explainable and accurate.
  
%   We experiment with the BERT model on the dataset, and calculate the explainability and bias metrics as well. We then have a deep dive into misclassified texts and the social implications of the same. We go beyond the scope of the original paper to analyse the misclassified examples based on target community, and provide some extra visualizations to see the labels that are misclassified. Lastly, we provide some well-thought ways for possible improvements on the limitations presented. 
  
\end{abstract}

\section{Introduction}

Societal and cultural bias has an extremely negative effect on humanity. It causes divide and sparks hatred among factions. It is thus one of the goals of every natural language researcher to ensure that such bias does not creep into his/her models and to take the necessary steps to rectify the same, in case it does happen. There are a very few datasets out there that actually try to aim for explanability of hate speech models and why they predict as they do. The topic of hate speech detection is a part of a bigger subset of problems like toxic span detection, offensive language detection, implicit bias recognition, among others.  

\subsection{Problem Statement - What is hate speech?}

Hate speech is defined as a group of words or phrases that are targeted towards a certain group of people and are derogatory in nature. Hate speech can fall under many categories - against certain sexual orientation (homophobic), against certain religions (islamophobic), against certain races (racial slur), etc. \\\\
\textbf{Hate Speech detection} is the task of detecting such sentiment from text. The task is to distinguish between normal language, offensive language and hate speech. While the distinction with the first category is fairly straightforward, the distinction among the latter 2 categories is non-trivial. Additionally, in order to make the task more human-detection-like, we also include the target community and span of words that cause the prediction. This makes our problem statement more inclined towards explanability - helping us answer as to why the deep models predict the way they do. 

\subsection{Our contributions}
We tackle the problem of hate speech detection by using the world's first rationale-based dataset on this topic - HateXplain. Our BERT implementations work well with the plethora of metrics available at our disposal in all three categories - accuracy, explainability and bias. Specifically we do the following in our work:

\begin{itemize}
    \item $\lambda$ (attention rational importance) tuning on the BERT model. The hypothesis of the authors is tested to see if more attention to rationales improves the performance.
    \item We introduce new methods to take the rationale attention itself - conservative attention (all annotators agree), lenient attention (anyone agrees), and normal attention (mean attention). 
    \item We also use masking of the target community names to improve on the bias metrics and meet the expectations of our hypothesis, albeit at a very minor cost of overall accuracy (which was also expected due to loss of information)
    \item Finally we conclude with an extensive error analysis that is interesting in terms of the improvements on different fronts. We visually see the improvements in class-wise prediction and a reduction in marginal community missclassification.
\end{itemize}

\section{Literature Survey - Related Works}

Hate speech and toxic span prediction has been a relevant topic in the last decade. There have been multiple attempts at the problem statement, with a wide range of methods.\\\\
A survey paper by \cite{schmidt2017survey}  highlights how hate speech detection has transitioned in time. It explains that bag of words fails to generalise as it depends on word occurrences instead of embeddings/features. They analyze the role of knowledge-based features and meta-information into the models, and urge for datasets involving granular classification, and focus on aspects other than the class.\\\\
\cite{burnap2016us} tried using bag of words combined with type dependencies to extract features on text. this was followed by a variety of statistical ML methods like RBF kernel SVMs and Random Forest.\\\\
There has also been some interesting work by \cite{ribeiro2018characterizing} on a more user-centric view of hate speech. They concluded that hate speech practicing Twitter users are more impulsive in tweeting at shorter intervals, and have a peculiar retweet graph and vocabulary (word pattern). \cite{lexico_semantic_features} tried using feature engineering to augment classifiers and compare their performance with each other. Their experimentation showed that adding lexico-semantic features like boolean hate word vectors, sentiment vector embeddings, pre-trained skip-gram word embeddings and n-gram embeddings boost performance of simple LSTMs, and SVM based classifiers. \\\\
A very recent work by \cite{bert-toxic} breaks down the words into tokens using the BERT tokenizer. The characters associated with the original word formulation are noted and the model goes on to predict the tokens as toxic or normal. Further, as a post processing step, all characters between two consecutive toxic tokens are labelled as toxic too. The rationale behind doing this is to include delimiters and characters that occur between two toxic tokens as also toxic in context. This helps detect toxic phrases, and refines toxic span boundaries to boosts test accuracy by over 4 percent.\\\\
Two works by  the same group of researchers - \cite{mudes} and \cite{wlv_toxic_spans} - explore and highlight the use of neural transformers in hate speech and toxic span detection. Their architecture has a language modelling transformer, and a second token predicting transformer. They try experimenting with Conditional random field layers between two transformers, only to find that it is not giving the boost that they expected. They introduce the MUDES framework to try different BERT architectures like RoBERTa and XLNet (\cite{yang2019xlnet}) to boost accuracy. They are successful in boosting accuracy this way by use of improved BERT models. \\\\
An interesting application of hate speech detection is movie subtitle classification and automated movie ratings based on the same. \cite{movieshateful} worked on this area and created a new dataset for the same task with 6 movies. They tested Bag of Words, Bi-directional LSTMs, and BERT models on their datasets. They also tested BERT models pre-trained on HateXplain to compare performance and enhance learning. The qualitative analysis for the best-performance BERT model shows that it is unsuccessful at detecting sarcastic hate speech that does not contain the classic hate words, and also misclassified some normal text like 'Black power' to be hate speech.

\section{About the dataset - HateXplain}

\subsection{Tackling multiple aspects}
The HateXplain dataset by \cite{HateXplain} is unique in 3 facets or factors. 

\begin{enumerate}
    \item It focuses on the category of speech - normal, offensive or hate speech. 
    \item It assigns the targeted community to the post. This community is based on religion, sex, caste, race, geography etc. 
    \item It associates word and phrase level spans of the text with the prediction.
\end{enumerate}

 These characteristics are in line with the growing trend and need to make deep learning models more interpretable and explainable. Using the rationale in combination with the prediction class and target community is also expected to make the model predictions more human-like.

\subsection{Differentiating between offensive and hate speech}
Another unique aspect  about the dataset is the distinction it makes between offensive and hate speech by using the standard set by  \cite{davidson2017automated}. This is extremely relevant in today's time and age when there is immense offensive content on the internet, not all of which comes under the category of hate speech because it is not targeted to be derogatory towards a particular group. Examples of such instances cited by Davidson et. al. in \cite{davidson2017automated} include the use of 'n*gga' by the African American Community, the use of 'b*tch' and 'wh*re' in rap lyrics etc. \\\\
\textbf{Normal text often misclassified as offensive by BERT-like models - full of words that are usually offensive.}  \\
\textit{Anyone leaked the info on who at youtube is the white supremacist neonazi fiddling with the algo seems that should drop next right along with the metrics linking up the infosec talking heads with their harassment and nazi socks right guise.}

\subsection{Dataset annotation}
The annotation has been done using Amazon MTurk and the selected annotators are all given batch-wise posts to annotate in several stages as highlighted by \cite{HateXplain}. The target label as well as the targeted class are decided upon majority voting. The annotations are given by annotators as boolean vectors equal to the length of the sentence. They are averaged, and then a softmax is applied to make the probability uniform and normalized. This acts as a ground truth attention, deactivated if the label is 'normal' speech.

\begin{table*}[hbt!]

\begin{center}

\caption{Performance Metrics for diff. $\lambda$  values}

% \resizebox{\columnwidth}{!}{%
\begin{tabular}{||c c c c||} 
 \hline
 Model ($\lambda$) & Accuracy & F1 Score & AUROC \\ [0.5ex] 
 \hline\hline
 0.001 & 0.669 & 0.669 & 0.852 \\
 \hline
 1 & 0.686 & 0.677 & 0.841 \\
 \hline
 10 & 0.685 & 0.676 & 0.840  \\
 \hline
 100 & 0.689 & 0.682 & 0.851 \\
 \hline
%  Original paper & 0.698 & 0.687 & 0.851 \\
%  \hline
\end{tabular}
% }
\label{tab:perf}
\end{center}
\end{table*}

\begin{table*}[hbt!]
\begin{center}
\caption{Bias Metrics for different $\lambda$ values}
% \resizebox{\columnwidth}{!}{%
\begin{tabular}{||c c c c||} 
 \hline
 Model ($\lambda$) & Subgroup AUC & BPSN & BNSP \\ [0.5ex] 
 \hline\hline
  0.001 & 0.814 & 0.764 & 0.79 \\
  \hline
  1 & 0.788 & 0.705 & 0.799 \\
  \hline
  10 & 0.765 & 0.707 & 0.788 \\
  \hline
  100 & 0.761 & 0.713 & 0.792 \\ 
 \hline
%  Original paper & 0.807 & 0.745 & 0.763 \\
%  \hline
\end{tabular}
% }
\label{tab:bias}
\end{center}
\end{table*}

\begin{table*}[hbt!]
\begin{center}
\caption{Explainability Metrics for different attention $\lambda$  values}
\begin{tabular}{||c c c c c c||} 
 \hline
 Model ($\lambda$) & IOU F1 & Token F1 & AUPRC & Comprehensiveness & Sufficiency \\ [0.5ex] 
 \hline\hline
 0.001 & 0.137 & 0.445 & 0.685 & 0.415 & 0.122 \\
 \hline
 1 & 0.144 & 0.554 & 0.84 & 0.607 & 0.135 \\
 \hline
 10 & 0.141 & 0.553 & 0.842 & 0.592 & 0.112 \\
 \hline
 100 & 0.142 & 0.567 & 0.868 & 0.585 & 0.07 \\ 
 \hline
%  Original paper & 0.120 & 0.411 & 0.626 & 0.424 & 0.160  \\
%  \hline
\end{tabular}
\label{tab:expl}
\end{center}
\end{table*}

\section{Performance, Explainability and Bias Metrics}

\subsection{Performance Metrics}
Comprises of metrics such as Accuracy. AUROC and F1-score. These reflect the statistical traditional performance of  the model over the data. 

\subsection{Bias Metrics}
The HateXplain paper explains that there need to be metrics to measure the unintended bias on certain words and classes. As per the standard and metrics set by \cite{bias_metrics}, we use multiple metrics. \\\\
A higher value of \textbf{Subgroup AUC} means that the model is doing a good job at distinguishing the toxic and normal posts specific to the community. A higher value of \textbf{Background Positive Subgroup Negative (BPSN) AUC.} means that a model is less likely to confuse
between the normal post that mentions the community
with a toxic post that does not. A higher value of \textbf{Background Negative Subgroup Positive (BNSP) AUC} means that the model is less likely to confuse between a toxic post that mentions the community with a normal post without one.

\subsection{Explainability metrics}
Plausibility refers to how convincing the interpretation is to humans, while faithfulness refers to how accurately it reflects the true reasoning process of the model. To measure the \textbf{plausibility}, we consider metrics for both discrete and soft selection - the IOU F1- Score and token F1-Score metric for the discrete case, and the AUPRC score for soft token selection. The inspiration is taken from \cite{deyoung2019eraser}. A high value of \textbf{comprehensiveness} implies that the rationales were influential in the prediction. \textbf{Sufficiency} measures the degree to which extracted rationales are adequate for a model to make a prediction.

\begin{table*}[hbt!]
\begin{center}
\caption{Performance metrics on Normal, Conservative and Lenient Ground Truth Attention}
\begin{tabular}{||l|l|l|l|l||}
\hline
Models & Lambda & Accuracy & F-score & ROCAUC  \\
\hline\hline
Baseline (Normal Attention) & 100    & 0.689    & 0.682   & 0.851     \\
\hline
Normal Attention            & 1      & 0.686    & 0.677   & 0.841      \\
\hline
Conservative Attention      & 100    & 0.700    & 0.687   & 0.855   \\
\hline
Conservative Attention      & 1      & 0.690    & 0.678   & 0.842  \\
\hline
Lenient Attention           & 100    & 0.680    & 0.666   & 0.844   \\
\hline
Lenient Attention           & 1      & 0.680    & 0.670   & 0.838   \\
\hline
\end{tabular}
\label{tab:perf_attention}
\end{center}

\end{table*}

\begin{table*}[hbt!]
\begin{center}
{
\caption{Bias and Explainability Metrics with Conservative and Lenient Attentions}
\begin{adjustbox}{width=\textwidth}
\begin{tabular}{||c|c|ccc|ccccc||}
\hline
\multirow{2}{*}{Model}                                                               & \multirow{2}{*}{Lambda} & \multicolumn{3}{c|}{Bias Metrics}                                      & \multicolumn{5}{c|}{Explainability Metrics}                                                                                                     \\ \cline{3-10} 
        & & \multicolumn{1}{c|}{Subgroup AUC} & \multicolumn{1}{c|}{BPSN}  & BNSP  & \multicolumn{1}{c|}{IOU F1} & \multicolumn{1}{c|}{Token F1} & \multicolumn{1}{c|}{AUPRC} & \multicolumn{1}{c|}{Comprehensiveness} & Sufficiency \\ \hline
\begin{tabular}[c]{@{}c@{}}Original Implementation\\ (Normal Attention)\end{tabular} & 100                     & \multicolumn{1}{c|}{0.761}        & \multicolumn{1}{c|}{0.713} & 0.792 & \multicolumn{1}{c|}{0.120}  & \multicolumn{1}{c|}{0.411}    & \multicolumn{1}{c|}{0.626} & \multicolumn{1}{c|}{0.424}             & 0.160       \\ \hline
Lenient Attention                                                                    & 100                     & \multicolumn{1}{c|}{0.794}        & \multicolumn{1}{c|}{0.722} & 0.788 & \multicolumn{1}{c|}{0.144}       & \multicolumn{1}{c|}{0.556}         & \multicolumn{1}{c|}{0.858}      & \multicolumn{1}{c|}{0.497}                  &      0.032       \\ \hline

Conservative Attention                                                               & 100                     & \multicolumn{1}{c|}{0.796}        & \multicolumn{1}{c|}{0.741} & 0.796 & \multicolumn{1}{c|}{0.139}  & \multicolumn{1}{c|}{0.535}    & \multicolumn{1}{c|}{0.819} & \multicolumn{1}{c|}{0.499}             & 0.0754      \\ \hline
Conservative Attention                                                               & 1                       & \multicolumn{1}{c|}{0.777}        & \multicolumn{1}{c|}{0.691} & 0.786 & \multicolumn{1}{c|}{0.138}  & \multicolumn{1}{c|}{0.540}    & \multicolumn{1}{c|}{0.832} & \multicolumn{1}{c|}{0.591}             & 0.129       \\ \hline
\end{tabular}
\label{tab:bias_expl_attention}
\end{adjustbox}
}

\end{center}
\end{table*}

\section{Experimentation, Tuning and Novel Contributions}

In this section, we cover our experiments, baselines, and performance on the HateXplain dataset and also highlight the multiple novel ideas that we try to incorporate to increase model accuracy and explanability. We try to provide the hypothesis/intuition behind our experimentation, and explain the results that follow.

\subsection{Model Deep-Dive - Using BERT}
The model is interpreted in two steps. Firstly, we calculate $ L_{pred} $ that is the loss on the predicted class. Secondly, the BERT model has an extra fully connected layer at the end. We try to match the attention values of the CLS token in this last layer to the ground truth attention and use a cross entropy loss to do so. This gives us $L_{att}$. The $\lambda$ value controls the trade off between the two.
$$L_{total} = L_{pred} + \lambda L_{att} $$

\subsection{Hyperparameter tuning}
Without a doubt, the most important hyperparameter in our BERT model was the $\lambda$ value, which regularizes how much effect the attention loss has on the total loss. We have reported the efficacy of tuning the hyperparameter $\lambda$ for three different classes of metrics:
\begin{enumerate}
    \item Performance Metrics
    \item Bias Metrics
    \item Explainability Metrics
\end{enumerate}

We observe that the performance of the model improves (in general) with increase in $\lambda$. These trends are observable from Table \ref{tab:perf}. This was in sync with our hypothesis, because hate speech detection is a contextual task, that should improve when relevant words are better attended to. In fact, this was the reason why bag of words-like models were not performing very well on this task. 
Table \ref{tab:bias} reports the values of the bias metrics (Subgroup AUC, BPSN and BNSP) for different values of $\lambda$. We observed that all three metrics were higher at lower values of $\lambda$, as opposed to the performance metrics. 
Finally, we have tabulated the effect of $\lambda$ on the explainability of the model in Table \ref{tab:expl}. We observed that the with the exception of Comprehensiveness and Sufficiency, the Explainability of the model is generally better at higher values of $\lambda$. However, in case of Comprehensiveness and Sufficiency, the best values were obtained for $\lambda = 1$.

\subsection{Modifying the Ground Truth Attention}
Currently, the ground truth attention is calculated by taking an average of the attention from all three annotators, followed by a softmax function to convert it to probabilities. This gives equal importance to all annotators. Since different annotators come from different backgrounds and perceptions, a sensitive task like hate speech detection can benefit from a non-equal importance. We thus divide experimentation into 3 approaches:\\\\
\textbf{Normal Attention:} This is the original papers method as explained above.\\\\
\textbf{Conservative Attention:} We attend to a word only if all annotators think it is important. This is analogous to an AND operator. We hypothesize that this change should perform better as the attended will be stronger than the normal mode.\\\\
\textbf{Lenient Attention:} We give importance to a word if any annotator thinks it is important. This is analogous to an OR operator. We hypothesize that this would be similar or lesser than the normal attention as we attend to all words with greater leniency.\\\\
\textbf{Improvements and Intuition:} We notice from Table \ref{tab:perf_attention} that we indeed gain around 1.1\% accuracy by using conservative attention, thus confirming our hypothesis. The intuition behind this increase is that the BERT model's last layer attends to more strongly prediction-correlated words when we we attend only to words marked by all annotators - thus removing bias from the annotations and model. \\\\
Clearly, from Table \ref{tab:bias_expl_attention} we can see that we were correct. We have relatively stable and high values from bias metrics in case of conservative attention, and massive improvements in case of explainability metrics, with the highest being for conservative attention with $\lambda =1$. Another interesting observation is the fact that lenient attention performs better on explanability metrics, because the active attention weights most definitely capture all words important for prediction. Hence, despite not seeing an accuracy boost like conservative attention, lenient attention weights contribute more to human interpretability, as reflected by high plausibility values.

% \begin{figure*}[hbt!]
% \centering
% \includegraphics[width=0.75\textwidth]{target comm.png}
% \caption{Target Community wise improvements}
% \label{missclass}
% \end{figure*}

\begin{table*}[t]
\begin{center}
\caption{Performance metrics on Masking Community Words}
\begin{tabular}{||l|l|l|l||}
\hline
Lambda & Accuracy & F-score & ROCAUC  \\
\hline\hline
100    &  0.694  & 0.689  &  0.847    \\
\hline
1   & 0.67   & 0.666   & 0.840  \\
\hline

\end{tabular}
\label{tab:perf_masking}
\end{center}
\end{table*}

\begin{table*}[hbt!]
\begin{center}
\caption{Bias and Explainability Metrics with Masking Community Words}
\begin{adjustbox}{width=\textwidth}

\begin{tabular}{||c|c|c|c|c|c|c|c|c||}
\hline
Lambda & \multicolumn{3}{c|}{Bias Metrics} & \multicolumn{5}{c|}{Explainability Metrics}                  \\
\hline
       & Subgroup AUC   & BPSN   & BNSP   & IOU F1 & Token F1 & AUPRC & Comprehensiveness & Sufficiency \\
       \hline
100    & 0.730          & 0.703  & 0.803  & 0.143  & 0.562    & 0.860 & 0.570             & 0.0865      \\
\hline
1      & 0.675          & 0.655  & 0.788  & 0.142  & 0.543    & 0.824 & 0.558             & 0.104      \\
\hline
\end{tabular}

\end{adjustbox}
\label{tab:bias_expl_masking}
\end{center}
\end{table*}

\begin{figure*}[!ht]
\centering
\includegraphics[width=0.75\textwidth]{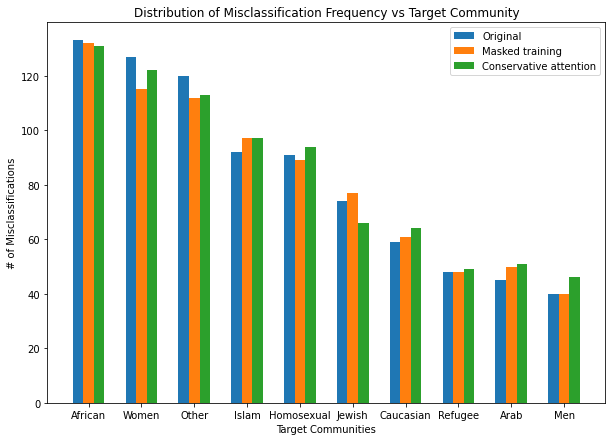}
\caption{Target Community wise misclassifications}
\label{missclass}
\end{figure*}

\begin{figure*}[h]
\centering
\includegraphics[width=\textwidth]{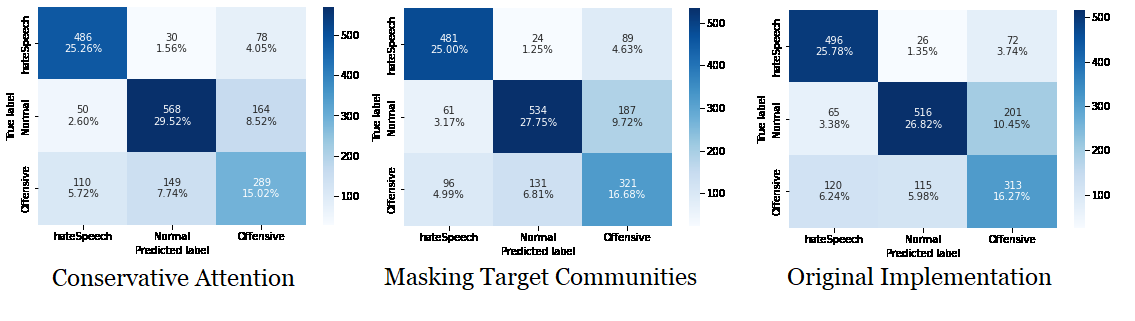}
\caption{Confusion Matrix for the 3 models}
\label{fig2}
\end{figure*}

\begin{figure*}[h]
\centering
\includegraphics[width=0.75 \textwidth]{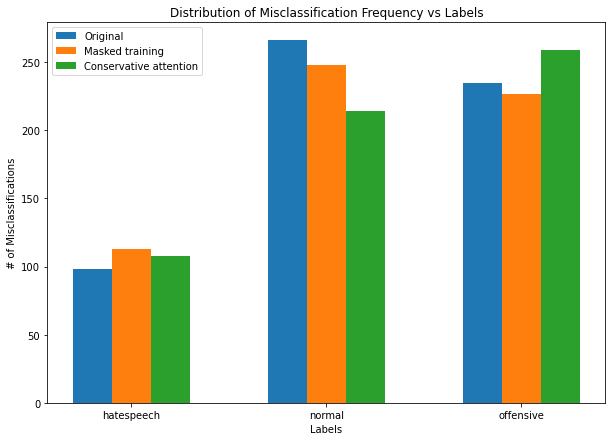}
\caption{Class-wise misclassifications}
\label{fig3}
\end{figure*}

\subsection{Masking Community Names}

We noticed in the original implementation that there are many instances of erroneous classification whenever the model comes across offensive words. These are the cases where the model classified the text as \textit{hatespeech} or \textit{offensive} just because they found a word belonging to a targeted community even though the statement did not convey any hate and was in fact a normal sentence (eg. rap songs with the word n*gga). We have attempted to reduce these kind of misclassifications by masking certain key words from these targeted communities during the training of the model. The intuition behind this is that if the model tries to judge a sentence just by the context and not the community then it wouldn't normal texts with with certain key words as toxic during testing. Masking the community names/target names on one hand should give the model an independence to classify based on non-community words, but it will lead to loss of information. As is visible from Table \ref{tab:bias_expl_masking} and Table \ref{tab:perf_masking}, the former reason boosts our explainability and bias metrics, while the latter slightly reduces our accuracy. Overall, the model performed as expected and was indeed successful in removing bias and increasing explainability. 

We found that masking target communities in the training data, despite increasing the accuracy slightly, leads to a reduction in the bias. We believe this is because the model might become more susceptible to classifying normal sentences as toxic, since it can be triggered by rationales such as "get rid of", "erased", "absolutely useless". While such terms - when used in the context of a target community - can constitute hatespeech, they can also be classified as normal speech if one were to talk about "getting rid of" trash. Another example would be [UNK] is extremely irritating, we must get rid of them. The same sentence would be considered toxic if the [UNK] token were to be a target community such as "Jews", "Muslims", "Blacks", etc., but would be a normal sentence if [UNK] was "Advertisements","Mosquitoes", etc. 
Given that Subgroup AUC, BNSP and BPSN are based on how well the model is able to identify toxic comments from a subgroup as compared to the background (all other communities combined), we believe these metrics may be negatively impacted since the model may incorrectly label normal sentences as toxic (more false positives).

\section{Error Analysis and Improvements}

It is not easy to understand how our modifications actually change the results of the model based on these metrics alone. To further consolidate our findings, we qualitatively explore the distribution of the misclassifications from each model to obtain more insights into the workings of the model.\\\\
For all analysis we have considered 3 models: the original Bert-HateXplain model, and the 2 modified models that gave us an increase in performance(Conservative Attention and Masked Training). In all our experiments, the best performance was obtained when $\lambda$ was 100 and hence that has been considered for comparisons and inference.

\subsection{Improvements in classification of marginalised classes}
 In order to understand the impact on community-specific performance, we plot the top 10 most misclassified communities in Figure \ref{missclass}. We see significant improvements and progress in reducing misclassifications on highly marginalised and targeted communities using both masking and conservative attentions. As is clearly visible from Figure \ref{missclass}, both of our proposed novelties reduce the misclassifications on highly missclassified and marginalised/targeted communities like African, women, and miscellaneous (other). We also observe some mixed trends in communities like Jewish, where conservative attention does very well due to strict control over rationales, but community masking does not (due to loss of information as highlighted before).

\subsection{Improvements in class-wise classification}
From Figure \ref{fig2} we can see that conservative attention is able to effectively tackle normal speech that was earlier classified as hate speech or offensive speech, because of its inherent stricter and conservative nature. There is however, minor increase in offensive speech classified as normal speech too. Overall, the confusion matrix of conservative attention is better than the masking and original implementations. A cumulative visual of our observations is given in Figure \ref{fig3}, which clearly demonstrates that we have been successful in reducing the miss-classification of normal text as offensive or hate speech.

\section{Conclusion}

In this paper, we thoroughly analyzed hate speech detection using BERT models on the world's first 'rationale'-based hate speech detection dataset - HateXplain. We looked at joint prediction and rationale loss, with different weights given to the two losses during tuning. Further, we propose the use of 2 new rationale attention mechanisms - conservative and lenient attention, the former boosting accuracy. We also analyze masking the target community names to reduce the bias in the classification, which is successful in the said aim at a marginal loss of information (accuracy). We end with a deep dive into the intuitions behind each of the proposed methods, and leverage illustrative class-wise and community-wise graphs and missclassification frequencies to visually see the positive impact of our changes. 

\section{Github Repository for submission}
\href{https://github.com/sayani-kundu/11711-HateXplain}{Click here for the Github Repo}\\\\
Else paste this link in the browser: https://github.com/sayani-kundu/11711-HateXplain

\bibliographystyle{abbrvnat}
\bibliography{sample.bib}

\end{document}